\documentclass[letterpaper, 10 pt, conference]{ieeeconf}  % Comment this line out 
\IEEEoverridecommandlockouts                              % This command is only needed if 
\title{\LARGE \bf
KDMOS:Knowledge Distillation for Motion Segmentation
}

\author{
    Chunyu~Cao,
    Jintao~Cheng,
    Zeyu~Chen,
    Linfan~Zhan,
    Rui~Fan,
    Zhijian~He,
    Xiaoyu~Tang\textsuperscript{*}% <-this % stops a space
    \thanks{(\textit{Corresponding author: Xiaoyu Tang}).}
    \thanks{Chunyu Cao, Jintao Cheng, Linfan Zhan, Zeyu Chen, and Xiaoyu Tang are with the School of Electronic and Information Engineering, South China Normal University, Foshan 528225, China. {\tt\small tangxy@scnu.edu.cn}}
    \thanks{Rui Fan is with the College of Electronics \& Information Engineering, Shanghai Research Institute for Intelligent Autonomous Systems, the State Key Laboratory of Intelligent Autonomous Systems, and Frontiers Science Center for Intelligent Autonomous Systems, Tongji University, Shanghai 201804, China. {\tt\small rui.fan@ieee.org}}
    \thanks{Zhijian He is with the College of Big Data and Internet, Shenzhen Technology University, Shenzhen, China. {\tt\small hezhijian@sztu.edu.cn}} % <- This right bracket was missing
}

\usepackage{amsmath}
\usepackage{bm}
\usepackage{cite}
\usepackage{graphicx}
\usepackage{amsfonts}
\usepackage{amsmath,amsfonts}
\usepackage{algorithmic}
\usepackage{algorithm}
\usepackage{array}
\usepackage[caption=false,font=normalsize,labelfont=sf,textfont=sf]{subfig}
\usepackage{textcomp}
\usepackage{stfloats}
\usepackage{url}
\usepackage{verbatim}
\usepackage{graphicx}
\usepackage{cite}
\usepackage{booktabs}
\usepackage{tabularx}
\usepackage{titlesec}
\usepackage{hyperref}
\usepackage{threeparttable}
\usepackage[switch]{lineno}

\begin{document}

\maketitle
\thispagestyle{empty}
\pagestyle{empty}

%%%%%%%%%%%%%%%%%%%%%%%%%%%%%%%%%%%%%%%%%%%%%%%%%%%%%%%%%%%%%%%%%%%%%%%%%%%%%%%%
\begin{abstract}

Motion Object Segmentation (MOS) is crucial for autonomous driving, as it enhances localization, path planning, map construction, scene flow estimation, and future state prediction. While existing methods achieve strong performance, balancing accuracy and real-time inference remains a challenge. To address this, we propose a logits-based knowledge distillation framework for MOS, aiming to improve accuracy while maintaining real-time efficiency. Specifically, we adopt a Bird’s Eye View (BEV) projection-based model as the student and a non-projection model as the teacher. To handle the severe imbalance between moving and non-moving classes, we decouple them and apply tailored distillation strategies, allowing the teacher model to better learn key motion-related features. This approach significantly reduces false positives and false negatives. Additionally, we introduce dynamic upsampling, optimize the network architecture, and achieve a 7.69\% reduction in parameter count, mitigating overfitting. Our method achieves a notable IoU of 78.8\% on the hidden test set of the SemanticKITTI-MOS dataset and delivers competitive results on the Apollo dataset.The KDMOS implementation is available at \href{https://github.com/SCNU-RISLAB/KDMOS}{https://github.com/SCNU-RISLAB/KDMOS}.

\end{abstract}

%%%%%%%%%%%%%%%%%%%%%%%%%%%%%%%%%%%%%%%%%%%%%%%%%%%%%%%%%%%%%%%%%%%%%%%%%%%%%%%%
\section{INTRODUCTION}

Accurate separation of moving and static objects is crucial for efficient path planning and safe navigation in dynamic traffic environments \cite{1}. Moving Object Segmentation (MOS), particularly for pedestrians, cyclists, and vehicles, reduces system errors caused by dynamic objects and improves environmental perception accuracy \cite{2}.This technology is essential for reducing uncertainties in scene flow estimation \cite{3},\cite{4} and path planning \cite{5}, enabling autonomous systems to make precise and reliable decisions. MOS plays a key role in real-time obstacle detection and adaptive environmental perception, making it integral to autonomous driving technology.

\begin{figure}
    \centering
    \includegraphics[width=1\linewidth]{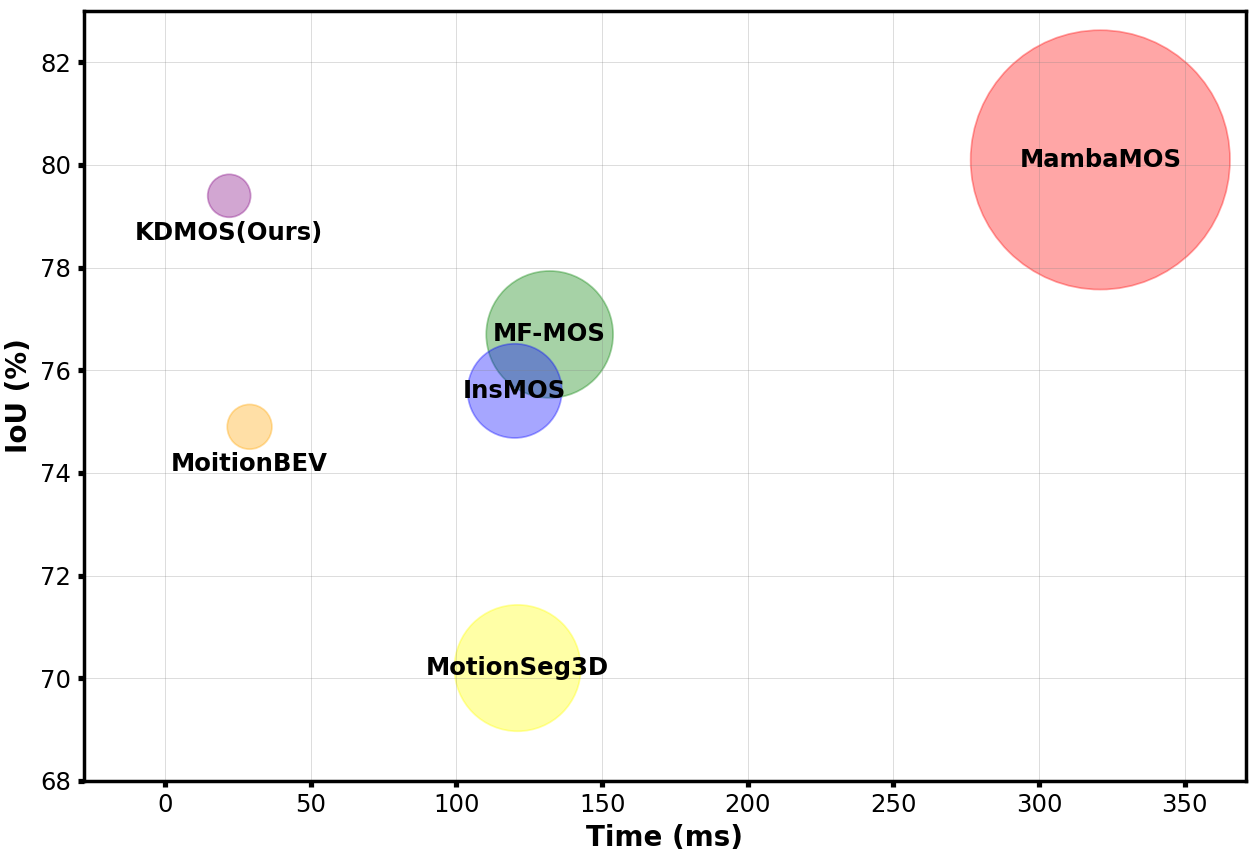}
    \caption{The performance, inference speed, and parameter size of different MOS models on the SemanticKITTI-MOS test set are compared. The size of the circles represents the parameter size. Our knowledge distillation-based KDMOS not only outperforms MotionBEV, MF-MOS, and InsMOS in performance but also maintains a very low computational cost.}
    \label{fig:1}
\end{figure}

For the MOS task, existing solutions can be categorized into projection-based \cite{2,6,7,8} and non-projection-based methods \cite{9,10,11}. Projection-based methods lose geometric information when mapping results back to the 3D point cloud space, limiting their performance. To address this, \cite{7} proposed a two-stage approach using 3D sparse convolution to fuse information from the projection map and point cloud, reducing back-projection loss and improving accuracy.However, this approach is computationally expensive, making it challenging to balance accuracy and inference speed. Likewise, non-projection-based methods \cite{9,10,11} face the same issue. Among them, MambaMOS\cite{11} achieves state-of-the-art (SoTA) performance in the MOS task, but its long processing time limits real-time applicability (see Fig.~\ref{fig:1}).

Knowledge distillation (KD) \cite{12} is an effective model compression technique that addresses real-time performance issues. Previous distillation algorithms \cite{13,21} have been successfully applied to LiDAR semantic segmentation, achieving remarkable results. However, most of these methods focus on model compression \cite{13} or improving feature extraction \cite{yan20222dpass}, with few dedicated to generalizable knowledge distillation methods for MOS tasks.

Based on the most intuitive observation that balancing accuracy and real-time performance is fundamental to solving the MOS problem, we propose KDMOS. The core idea of KDMOS is to compress the large model and balance inference speed and accuracy. We propose Weighted Decoupled Class Distillation (WDCD), a logit-based distillation method. Following the DKD approach \cite{14}, we first decouple traditional knowledge distillation into KL losses for target and non-target classes.To mitigate the severe class imbalance in the moving category for MOS tasks and enhance distillation performance, we further decouple moving and non-moving classes, apply different distillation strategies to compute their respective losses, and assign appropriate weights via labels.This allows the student model to effectively learn critical information among potential moving object categories, significantly reducing false positives and missed detections (see Fig.~\ref{fig:5}). Moreover, it can be broadly applied to other MOS tasks (see Fig.~\ref{fig:6}). Additionally, we introduce dynamic upsampling in the network, achieving an inference speed of 40 Hz while maintaining a balance between accuracy and speed.

Extensive experiments demonstrate the superiority of our design. In summary, the main contributions of this paper are as follows:
\begin{itemize}
    \item We propose a general distillation framework for the MOS task, effectively balancing real-time performance and accuracy (see Fig.~\ref{fig:1}). To the best of our knowledge, this is the first application of knowledge distillation to MOS.
        
    \item The KDMOS network architecture is improved by introducing the Dysample offset, reducing model complexity and mitigating overfitting.
        
    \item Our method achieves competitive results on the SemanticKITTI \cite{6} and Apollo \cite{18} datasets, demonstrating its superior performance and robustness.

\end{itemize}

\section{RELATED WORK}

\subsection{Knowledge Distillation}

Knowledge distillation (KD) was first introduced by Hinton et al. \cite{12}. Its core idea is to bridge the performance gap between a complex teacher model and a lightweight student model by transferring the rich knowledge embedded in the teacher. Existing methods can be categorized into two types: distillation from logits \cite{12,14,19,20} and from intermediate features \cite{13,21}. Hou et al. \cite{13} proposed Point-to-Voxel Knowledge Distillation (PVKD), the first application of knowledge distillation to LiDAR semantic segmentation.They introduced a supervoxel partitioning method and designed a difficulty-aware sampling strategy. Feng et al. \cite{21} proposed a voxel-to-BEV projection knowledge distillation method, effectively mitigating information loss during projection.Borui et al. \cite{14} decouple the classical KD formula, splitting the traditional KD loss into two components that can be more effectively and flexibly utilized through appropriate combinations.
Although feature-based methods often achieve superior performance, they incur higher computational and storage costs and require more complex structures to align feature scales and network representations.
Compared to the simple and effective logit-based approach, it has weaker generalization ability and is less applicable to various downstream tasks.

\subsection{MOS Based on Deep Learning}

Recent methods tend to apply popular deep learning models and directly capture spatiotemporal features from data. Chen et al \cite{6} proposed the LMNet, which uses range residual  images as input and extracts temporal and spatiotemporal information through existing segmentation networks. Sun et al. \cite{7} introduced a dual-branch structure that separately processes range images and residual images, utilizing a motion-guided attention module for feature fusion. Cheng et al. \cite{2} suggested that residual images offer greater potential for motion information and proposed a motion-focused model with a dual-branch structure to achieve the decoupling of spatiotemporal information. Unlike RV projection, BEV projection represents point cloud features from a top-down view, preserving object scale consistency and improving interpretability and processing efficiency. MotionBEV \cite{8} projects to the polar BEV coordinate system and extracts motion features through height differences over temporal windows. CV-MOS\cite{tang2024cv} proposes a cross-view model that integrates RV and BEV perspectives to capture richer motion information. Non-projection-based methods directly operate on point clouds in 3D space. 4DMOS \cite{9} uses sparse four-dimensional convolution to jointly extract spatiotemporal features from input point cloud sequences and integrates predictions through a binary Bayesian filter, achieving good segmentation results. Mambamos \cite{11} considers temporal information as the dominant factor in determining motion, achieving a deeper level of coupling beyond simply connecting temporal and spatial information, thereby enhancing MOS performance. Therefore, previous methods have struggled to balance real-time performance and accuracy. To address this issue, this paper proposes KDMOS based on knowledge distillation, which distills a non-projection-based large model into a projection-based lightweight model through logits distillation.

\section{METHODOLOGY}

\begin{figure*}
    \centering
    \includegraphics[width=0.9\textwidth, height=7.2cm]{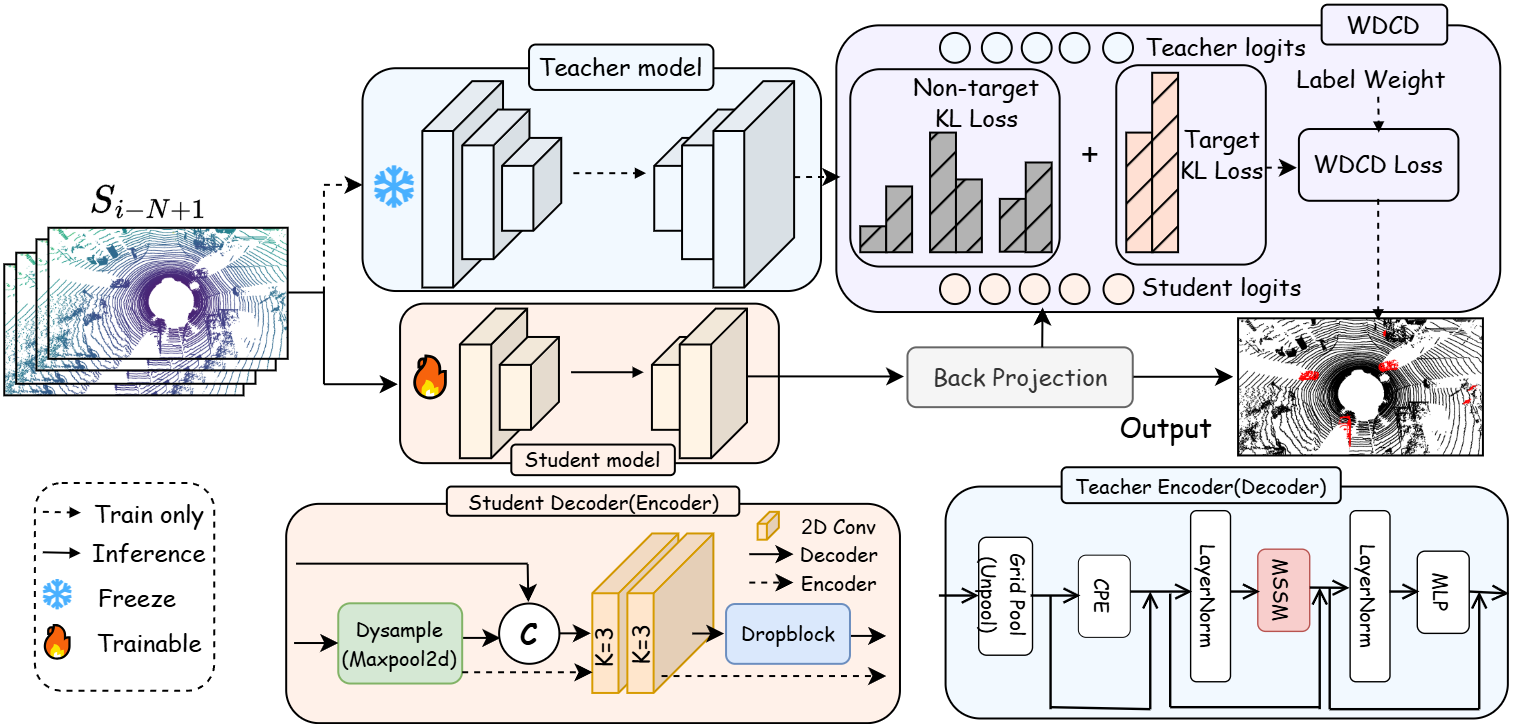}
    \caption{The KDMOS framework comprises three main components: the teacher model, the student model, and knowledge distillation. During training, the teacher model uses pre-trained weights and remains frozen, while the student model is trained from scratch, with its parameters continuously updated through WDCD and the model’s own loss. MSSM, proposed by MambaMOS\cite{11}, achieves deep coupling of temporal and spatial features.}
    \label{fig:2}
\end{figure*}

In this section, we present a detailed description of KDMOS, with its overall framework illustrated in Fig.~\ref{fig:2}. We start with data preprocessing, followed by an explanation of the KDMOS network architecture and WDKD. Finally, we provide an in-depth analysis of the loss function components.

\subsection{Input Representation} 

\textbf{Student Input Representation.} We employ an extreme bird's-eye view (BEV) for point cloud coordinate segmentation, a lightweight data representation obtained by projecting the 3D point cloud into 2D space. Following the setup from previous work \cite{8}, we project the LiDAR point cloud onto a BEV image. After obtaining the BEV images of the past N-1 consecutive frames, we align the past frames with the current frame's viewpoint using pose transformation \( T \in \mathbb{R}^{4 \times 4} \), resulting in the projected BEV images. Meanwhile, we maintain two adjacent time windows, \(Q_{1}\) and \(Q_{2}\), with equal lengths, and obtain the residual image by computing the height difference between the corresponding grids in the two time windows.
\begin{equation}
    \begin{aligned}
        Z_{(u,v),i} &= \{z_j \in p_j \mid p_j \in Q_{(u,v),i}, z_{\min} < z_j < z_{\max} \}, \\
        I_{(u,v),i} &= \max\{Z_{(u,v),i}\} - \min\{Z_{(u,v),i}\}.
    \end{aligned}
\end{equation}
where $p_j$ is a point within the temporal window $Q_{(u,v),i}$, represented as $[x_j, y_j, z_j, 1]^T$, and each pixel value $I_{(u,v),i}$ represents the height occupied by the $(u,v)_{th}$ grid in $Q_i$. Following MotionBEV \cite{8}, we restrict the z-axis range to $(z_{\text{min}}, z_{\text{max}}) = (-4, 2)$. Subsequently, we compute the residuals between the projected BEV images $I_1$ and $I_2$:

\begin{equation}
\resizebox{0.45\textwidth}{!}{$
\begin{aligned}
D_{(x, y),i}^0,D_{(x, y),i-1}^1,...,D_{(x, y),i-N_2+1}^{N_2-1}=I_{(x, y),1}-I_{(x, y),2},\\
D_{(x, y),i-N_2}^{N_2},D_{(x, y),i-N_2-1}^{N_2+1},...,D_{(x, y),i-N+1}^{N-1}=I_{(x, y),2}-I_{(x, y),1}
\end{aligned}
$}
\end{equation}
where \(D^{k}_{(u,v),i}\) represents the motion feature in the \((u,v)^{\text{th}}\) grid of the \(k^{\text{th}}\) channel for the \(i^{\text{th}}\) frame.

\textbf{Teacher Input Representation.}To align with the student's input, following the setup of previous work \cite{11}, we need to align the past frames with the current frame's viewpoint using the pose transformation matrix \( T \in \mathbb{R}^{4 \times 4} \) and convert the homogeneous coordinates into Cartesian coordinates, resulting in a sequence of N continuous 4D point cloud sets. To distinguish each scan within the 4D point cloud, we add the corresponding time step of each scan as an additional dimension of the point, obtaining a spatio-temporal point representation:
\begin{equation}
\begin{array}{c}
p'_i = \begin{bmatrix} x_i, y_i, z_i, t_i \end{bmatrix}^T, \\
S' = \{ S_0, S_{1 \rightarrow 0}, \dots, S_{t \rightarrow 0} \}
\end{array}
\end{equation}

Next, We follow the setup of MambaMOS \cite{11} to obtain sequences from unordered 4D point cloud sets.:

\begin{equation}
\begin{array}{c}
S'_i = \Psi(S'), \\
S' = \Psi^{-1}(S'_i)
\end{array}
\end{equation}

\subsection{Network Structure}
1)We propose a novel knowledge distillation network architecture, as shown in Fig.~\ref{fig:2}. The KDMOS network comprises three main components. First, the teacher model, where we select MambaMOS \cite{11} as the teacher, which deeply integrates temporal and spatial information, delivering strong performance but with a high computational cost.\begin{figure}[htbp] % h=here, t=top, b=bottom, p=page
    \centering
    \includegraphics[width=0.9\linewidth]{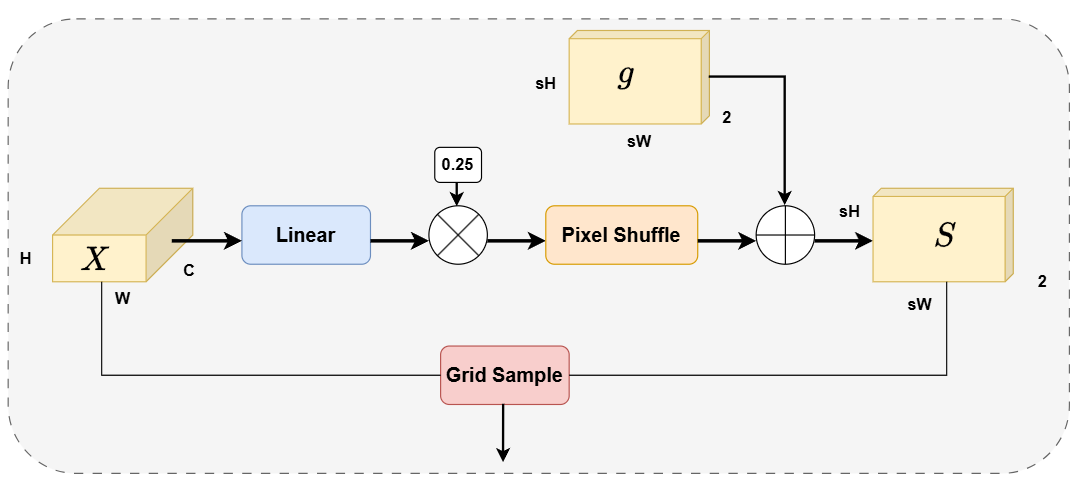}
    \caption{Structure of the Dysample module. The input feature and original grid are denoted by $X$  and $g$ , respectively.}
    \label{fig:3}
\end{figure} We select a BEV-based method \cite{8} as the student model. BEV projection provides a global top-down view, intuitively representing object distribution and relative positions in the scene while maintaining low computational complexity. The final component is the knowledge distillation model, the core of our framework. With WDCD (see Fig.~\ref{fig:4}), the student model learns category similarities and differences during training, enhancing performance without additional computational cost.

2)Feature Upsampling Module:To balance accuracy and inference speed, we optimize the upsampling module with the dynamic mechanism DySample \cite{22}. As shown in Fig.~\ref{fig:3}, the feature map is transformed through a linear layer, scaled by an offset factor to compute pixel displacement coordinates, and refined via pixel shuffle for upsampling. Displacement coordinates are added to the base grid for precise sampling. This approach replaces fixed convolution kernels with point-based sampling, reducing parameter count and model complexity, thereby lowering the risk of overfitting.

\subsection{Weighted Decoupled Class Distillation}
\label{1}

\begin{figure}
    \centering
    \includegraphics[width=1\linewidth]{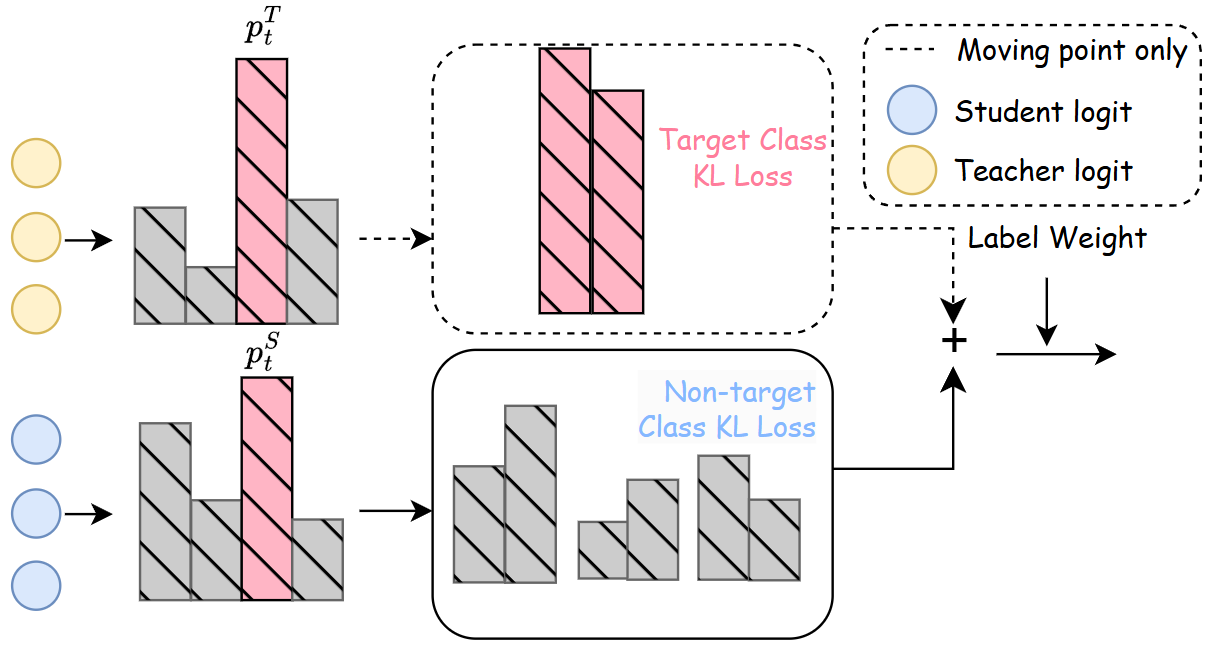}
    \caption{The structure of WDCD, where $p_t^T$ and $p_t^S$ represent the teacher's and student's probabilities for the target class, respectively.}
    \label{fig:4}
\end{figure}

WDCD is a key distillation module in our framework, as shown in Fig.~\ref{fig:4}. Most distillation methods \cite{13,21} rely on intermediate layer features from both the teacher and student networks.However, if the teacher and student architectures differ significantly, aligning feature scales and network capabilities requires more complex structures, making distillation more challenging and less effective. In contrast, logits-based distillation \cite{12,14} relies solely on the teacher’s final output, bypassing its internal structure. This approach is simpler to implement, more computationally efficient, and applicable to other MOS methods.

The logits contain relational information between categories, and to further leverage this information, unlike previous MOS methods \cite{2,6,7,8,9,10,11}, we divide the final predictions of points into four categories: unlabeled, static, movable, and moving.Specifically, for a training sample from the $t$-th class, we first compute the probabilities for the target and non-target classes for each point:

\begin{equation}
\label{pt}
    p_t = \frac{\exp(z_t)}{\sum_{j=1}^{4} \exp(z_j)}, \quad
    p_{\setminus t}  = \frac{\sum_{k=1, k\neq t}^{4} \exp(z_k)}{\sum_{j=1}^{4} \exp(z_j)}
\end{equation}
where $z_j$ represents the logit of the $j$-th class, and $p_t$ and $p_{\setminus t}$ denote the probabilities of target and non-target classes, respectively.We know that the distribution probabilities for each class can be represented as \( p = [p_1, p_2, p_3,p_4] \). Meanwhile, we define \( \hat{p} = [\hat{p}_1, \hat{p}_2, \hat{p}_3] \) to independently represent the probabilities for the non-target classes, excluding the target class (i.e., without incorporating its influence).Each element is computed as follows:

\begin{equation}
\label{pi}
    p_i = \frac{\exp(z_i)}{\sum_{j=1}^{4} \exp(z_j)}, \quad
    \hat{p}_i = \frac{\exp(z_i)}{\sum_{j=1, j \neq m}^{4} \exp(z_j)}.
\end{equation}

Specifically, we use the binary probability $b$ and the non-target probability $\hat{{p}}$ to represent knowledge distillation (KD), where $T$ and $S$ denote the teacher and student models, respectively.

\begin{equation}
\label{kd}
    \text{KD} = \text{KL}({p}^{T} \| {p}^{S}) 
    = p_t^{T} \log\left(\frac{p_t^{T}}{p_t^{S}}\right) + \sum_{i=1, i \neq t}^{C} p_i^{T} \log\left(\frac{p_i^{T}}{p_i^{S}}\right).
\end{equation}From Equations \ref{pt} and \ref{pi}, we obtain \(\hat{p}_i = p_i / p_{\backslash t}\). Thus, Equation \ref{kd} can be rewritten as:

\begin{align}
    \text{KD} &= p_t^{T} \log\left(\frac{p_t^{T}}{p_t^{S}}\right) 
    + p_t^{T} \log\left(\frac{p_{\setminus t}^{T}}{p_{\setminus t}^{S}}\right) \notag \\
    &\quad + p_{\setminus t}^{T} \sum_{i=1, i\neq t}^{C} \hat{p}_i^{T} \log\left(\frac{\hat{p}_i^{T}}{\hat{p}_i^{S}}\right) \notag \\
    &= \text{KL}(\bm{b}^{T} \| \bm{b}^{S}) + (1 - p_t^{T}) \text{KL}(\bm{\hat{p}}^{T} \| \bm{\hat{p}}^{S})
\end{align} where $\text{KL}(\bm{b}^{T} \| \bm{b}^{S})$ represents the similarity between the binary probabilities of the target class for the teacher and student models, while $\text{KL}(\bm{\hat{p}}^{T} \| \bm{\hat{p}}^{S})$ denotes the similarity between the probabilities of non-target classes for the teacher and student models.In the MOS task, the severe imbalance between moving and non-moving classes results in the number of non-moving points being approximately 400 times that of moving points. During training, the high accuracy of non-moving classes significantly reduces the effectiveness of $\text{KL}(\bm{b}^{T} | \bm{b}^{S})$, which may even become detrimental (as shown in Table \ref{tab:tckd&nckd}). Meanwhile, $\text{KL}(\bm{\hat{p}}^{T} \| \bm{\hat{p}}^{S})$ is also influenced by $p_t^{T}$, severely impairing the effectiveness of knowledge distillation.

Therefore, in the MOS task, we further decouple moving and non-moving classes. For moving classes, both losses are computed as usual, while for easily trainable non-moving classes, only $\text{KL}(\bm{\hat{p}}^{T} | \bm{\hat{p}}^{S})$ is applied.
We define this approach as Decoupled Class Distillation (DCD).
\begin{equation}
    \text{DCD} =
    \begin{cases} 
        \text{KL}(\bm{b}^{T} \| \bm{b}^{S}) + \beta\text{KL}(\bm{\hat{p}}^{T} \| \bm{\hat{p}}^{S}), & \text{class} = \text{moving} \\[8pt]
        \beta\text{KL}(\bm{\hat{p}}^{T} \| \bm{\hat{p}}^{S}), & \text{class} \neq \text{moving}
    \end{cases}
\end{equation}where \( \beta \) is the balancing coefficient. Based on label-assigned weighting, our final WDCD is formulated as:

\begin{equation}
\begin{array}{c}
W^i = \text{Content}[\text{label}], \\
\text{WDCD}=  \text{DCD} / W^i

\end{array}
\end{equation}Where Content represents the ratio of points from different categories in the \( i^{th}\) frame, and  label represents the ground truth labels of points in the \( i^{th}\) frame.

\subsection{Loss Function}
During the training process, the total loss function of this algorithm includes both the segmentation loss and the knowledge distillation loss:

\begin{equation}
\begin{aligned}
\mathcal{L}_{\text{Total}} &= \mathcal{L}_{\text{Student}} + \gamma \mathcal{L}_{\text{LDCD}},
\end{aligned}
\end{equation}where \( {L}_{\text{Student}} \) represents the segmentation loss generated by the student network,  \( {L}_{\text{LDCD}} \) is the losses of Knowledge Distillation and \( \gamma \) is the balancing coefficient.The student network loss consists of the cross-entropy loss \( {L}_{\text{wce}} \) and the Lovász-Softmax loss \( {L}_{\text{ls}} \). The loss function is defined as follows:

\begin{equation}
\begin{aligned}
\mathcal{L}_{\text{Student}} &= \mathcal{L}_{\text{wce}} + \mathcal{L}_{\text{ls}},
\end{aligned}
\end{equation}

\section{EXPERIMENTS}
\begin{table}[htbp]
\centering
\caption{COMPARISONS RESULT ON SEMANTICKITTI-MOS DATASET.}
\label{tab:1}
\begin{tabular}{l c c c}
\hline
\textbf{Methods} & \textbf{Publication} & \textbf{Validation(\%)} & \textbf{Test(\%)} \\ \hline
LMNet \cite{6} & ICRA 2021 & 63.8 & 60.5 \\ 
LiMoSeg \cite{mohapatra2021limoseg} & ArXiv 2021 & 52.6 & - \\ 
Cylinder3D \cite{zhu2021cylindrical}  & CVPR 2021 & 66.3 & 61.2 \\ 
RVMOS \cite{kim2022rvmos} & RAL 2022 & 71.2 & 74.7 \\ 
4DMOS \cite{9} & RAL 2022 & 77.2 & 65.2 \\ 
MotionSeg3D \cite{7} & IROS 2022 & 71.4 & 70.2 \\ 
InsMOS \cite{10} & IROS 2023 & 73.2 & 75.6 \\ 
MotionBEV \cite{8} & RAL 2023 & 76.5 & 75.8 \\ 
SSF-MOS \cite{4} & TIM 2024 & 70.1 & - \\ 
Mambamos \cite{11} & ACM MM 2024 & \textbf{82.3} & \textbf{80.1} \\ 
MF-MOS \cite{2} & ICRA 2024 & 76.1 & 76.7 \\ \hline
Ours & - & \underline{79.4} & \underline{78.8} \\ \hline
\end{tabular}
\end{table}\begin{table}[htbp]
\centering
\caption{Cross Val IoU performance of different methods on Apollo dataset.}
\label{tab:2}
\begin{tabular}{l c}
\hline
\textbf{Methods} & \textbf{Validation(\%)} \\ \hline
LMNet & 16.9 \\ 
MotionSeg3D & 7.5 \\ 
MotionBEV & 25.1 \\ 
MF-MOS & 49.9 \\ 
4DMOS & \textbf{73.1} \\ 
Ours & \underline{68.2} \\ \hline
\end{tabular}
\end{table}
In this section, we conduct extensive experiments to comprehensively evaluate KDMOS. Section IV-A introduces the datasets and evaluation metrics. In Section IV-B, we present quantitative and qualitative comparisons between our method and other state-of-the-art approaches on SemanticKITTI-MOS \cite{6} and Apollo \cite{18}. In Section IV-C, we conduct ablation experiments to evaluate the effectiveness of our knowledge distillation module. In Section IV-D, we perform a qualitative analysis to intuitively compare our algorithm with other SoTA approaches. Finally, in Section IV-E, we present the runtime performance of our method.

\subsection{Experiment Setups}
The SemanticKITTI-MOS dataset \cite{17} is the largest and most authoritative dataset for the MOS task, derived from the original SemanticKITTI dataset. We follow the setup from previous work \cite{8} for training, validation, and testing.
Additionally, to evaluate the generalization ability of our method across different environments, we tested it on another dataset, Apollo, following the setup by Chen et al. \cite{6}.

Our code is implemented in PyTorch. Experiments were conducted on a single NVIDIA RTX 4090 GPU with a batch size of 8. We trained KD-MOS for 150 epochs using Stochastic Gradient Descent (SGD) to minimize \( {L}_{\text{wce}} \), \( {L}_{\text{ls}} \), and \( {L}_{\text{WDCD}} \), with a momentum of 0.9 and a weight decay of 0.0001.
\begin{table}[htbp]
\centering % 表格居中 
\setlength{\tabcolsep}{10pt}
\caption{ABLATION EXPERIMENTS WITH PROPOSED MODULES.}
\resizebox{0.5\textwidth}{!}{ % 将表格调整为页面宽度的 50%
{\fontsize{10}{14}\selectfont % 将表格中的字体大小设置为较大
\begin{tabular}{llllll}
\hline
\textbf{Methods} & \multicolumn{2}{c}{\textbf{Component}} & \textbf{IoU(\%)} & \textbf{params(M)} \\ \hline
                 & \textbf{Dysample} & \textbf{WDCD}    &                   &                    \\ \hline
MotionBEV        &                   &                  & \multicolumn{1}{c}{76.5} & \multicolumn{1}{c}{4.42}    \\
KDMOS(i)         & \multicolumn{1}{c}{-} & \multicolumn{1}{c}{-} & \multicolumn{1}{c}{76.2} & \multicolumn{1}{c}{4.42}    \\ \hline
                 & \multicolumn{1}{c}{\textbf{\checkmark}} & \multicolumn{1}{c}{-} & \multicolumn{1}{c}{76.7} & \multicolumn{1}{c}{4.08}    \\
KDMOS(ii)        & \multicolumn{1}{c}{-} & \multicolumn{1}{c}{\textbf{\checkmark}} & \multicolumn{1}{c}{77.8} & \multicolumn{1}{c}{4.42}    \\
                 & \multicolumn{1}{c}{\textbf{\checkmark}} & \multicolumn{1}{c}{\textbf{\checkmark}} & \multicolumn{1}{c}{\textbf{79.4}} & \multicolumn{1}{c}{\textbf{4.08}}    \\ \hline
\end{tabular}
}
}
\label{ablation}
\end{table}\begin{figure}[htbp]
    \centering
    \includegraphics[width=0.7\linewidth, height=4.5cm]{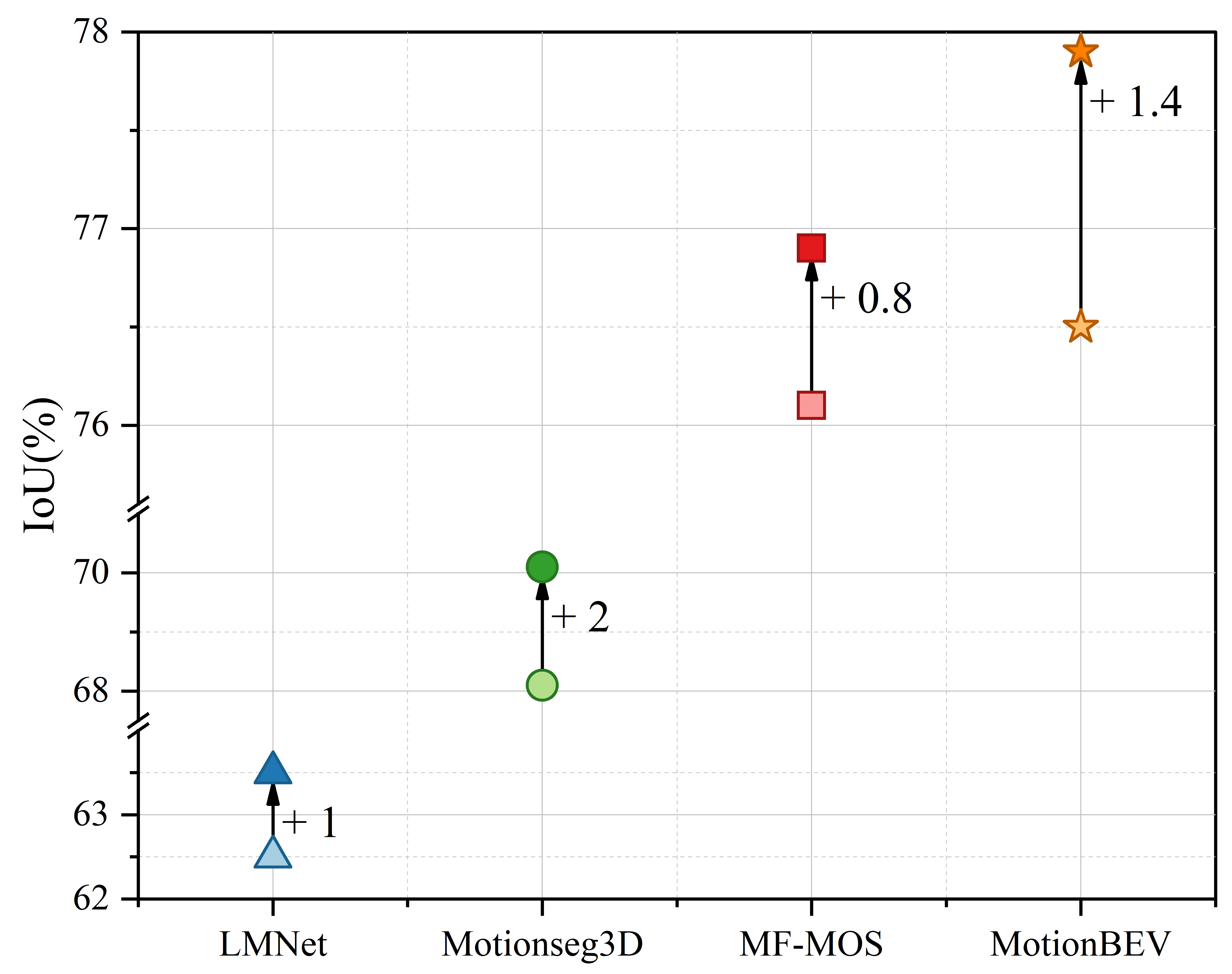}
    \caption{The performance of the proposed WDCD module on other MOS methods (IoU\%).}
    \label{fig:6}
\end{figure}
The initial learning rate was set to 0.005 and decayed by a factor of 0.99 after each epoch. The offset factor for DySample \( a \) was set to 0.25, and the weight of \( {L}_{\text{WDCD}} \) was set to 0.25. The teacher network used MambaMOS pre-trained weights with frozen parameters. The student network was trained from scratch without pre-trained weights, and training continued until the validation loss converged. We used the Jaccard Index, also known as the Intersection-over-Union (IoU) metric \cite{23}, to evaluate MOS performance for moving objects.

\subsection{Comparison with State-of-the-Art Methods}
First, we report the validation and test results on the SemanticKITTI-MOS dataset \cite{6} in Tab.~\ref{tab:1}. Competitive results were achieved on the validation set. For the test set, we submitted the moving object segmentation results to the benchmark server, which also demonstrated excellent performance, further confirming the model’s robustness.As shown in Tab.~\ref{tab:1}, our mIoU score is lower than that of MambaMOS \cite{11} in both the validation and test benchmarks. However, as shown in Tab.~\ref{tab:performance}, while non-projection-based methods achieve higher accuracy, they often struggle to balance accuracy and inference speed. In contrast, our method effectively combines the advantages of both projection-based and non-projection-based approaches, demonstrating superior overall performance.Specifically, compared to the baseline MotionBEV, the performance improved by 3.9\% on the validation set and 2.9\% on the test set.

To evaluate the generalization ability of our method in different environments, we also report the validation results on the Apollo dataset \cite{18} in the Tab.~\ref{tab:2}. Following the standard settings of previous methods \cite{10,chen2022automatic},the data in the Tab.~\ref{tab:2} does not use any domain adaptation techniques or retraining. Compared to other projection-based methods, our KDMOS performed the best. Although the results on the Apollo dataset are not as good as those of 4DMOS \cite{9}, as shown in the table, Our method demonstrates significantly faster inference speed and outperforms 4DMOS on large-scale datasets such as SemanticKITTI-MOS.
\begin{table}[htbp]
\centering
\caption{COMPARISONS OF DIFFERENT KD METHODS.}
\label{tab:8}
\begin{tabular}{l c c}
\hline
\textbf{Method} & \textbf{Publication} & \textbf{IoU(\%)} \\ \hline
KD \cite{12} & NIPS 2015 & 77.2 \\ 
DKD \cite{14} & CVPR 2022 & 77.6 \\ 
KD+Std \cite{sun2024logit} & CVPR 2024 & 77.4 \\ 
Ours & - & \textbf{79.4} \\ \hline
\end{tabular}
\end{table}\begin{table}[h]
    \centering
    \caption{Ablation of DKD modules on non-moving classes using the SemanticKITTI-MOS validation set}
    \label{tab:tckd&nckd}
    \begin{tabular}{cc|c}
        \hline
        \multicolumn{2}{c|}{\textbf{Component}} & \textbf{IoU(\%)} \\
        \textbf{NCKD} & \textbf{TCKD} &  \\
        \hline
        - & - & 76.5 \\
        - & \checkmark & 76.2 \\
        \checkmark & - & 77.3 \\
        \checkmark & \checkmark & 77.1 \\
        \hline
    \end{tabular}
\end{table}
\subsection{Ablation Experiment}

In this section, we conduct ablation experiments on the proposed KDMOS and its various components. All experiments are performed on the SemanticKITTI validation set (sequence 08). As shown in Tab.~\ref{ablation}. It is noteworthy that our proposed WDCD shows significant improvement (+1.3\% IoU) compared to MotionBEV without increasing the number of parameters. To further demonstrate the indispensability of each component, we conducted ablation experiments with different component combinations in Setting ii. Each proposed component consistently improves baseline performance to varying degrees.The last row shows that our complete KDMOS achieves the best performance, with a significant accuracy improvement (+2.9\% IoU) and a 7.69\% reduction in parameter count compared to the baseline.

\begin{figure*}
    \centering
    \includegraphics[width=1\textwidth, height=8cm]{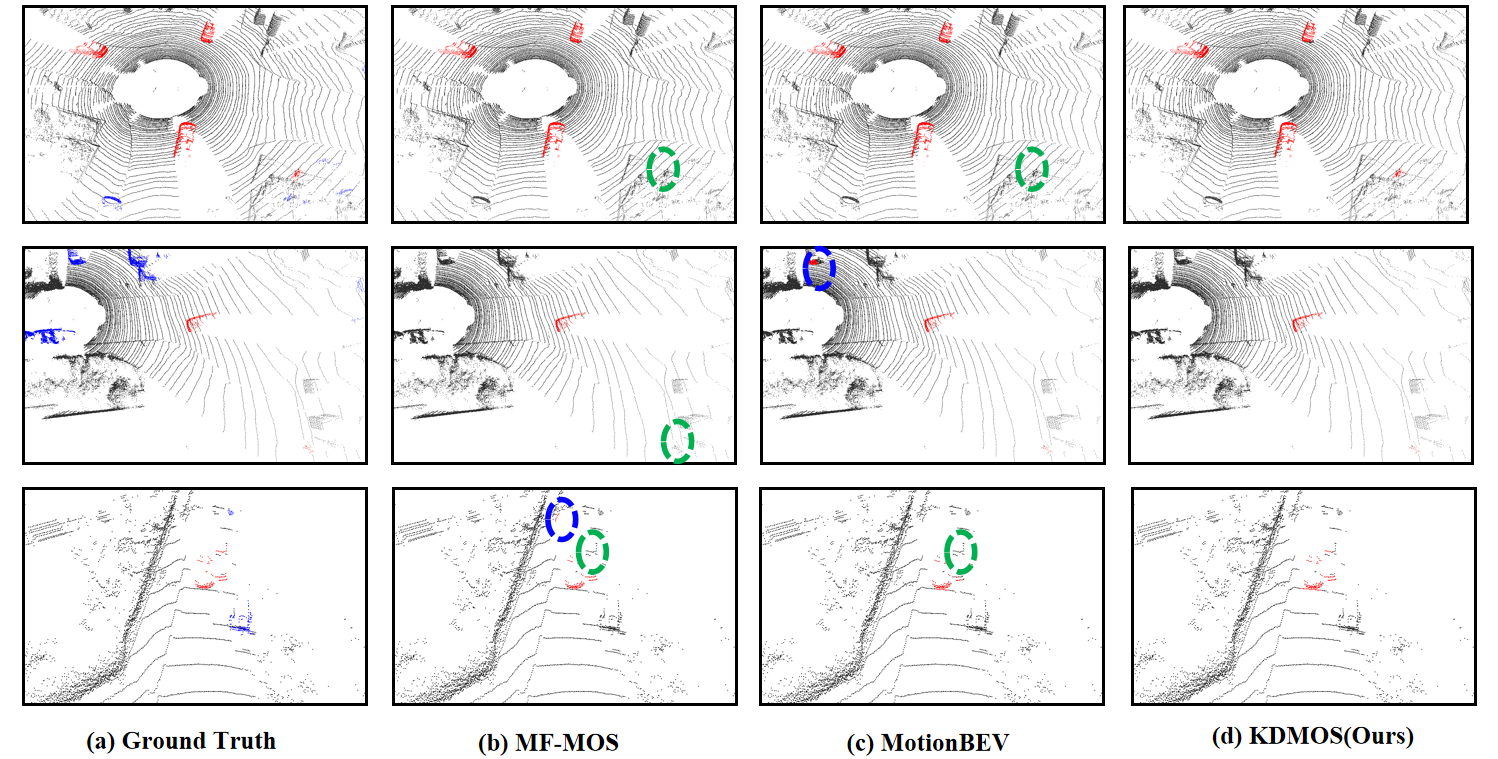}
    \caption{Qualitative results of LiDAR-MOS on the SemanticKITTI-MOS validation set using different methods. Green circles indicate false negatives, while blue circles indicate false positives.}
    \label{fig:5}
\end{figure*}\begin{table}[htbp]
\centering
\caption{COMPUTATION RESOURCE COMPARISON.}
\label{tab:performance}
\begin{tabular}{l c c c c c}
\hline
\textbf{Method} & \textbf{FPS} & \textbf{ms} & \textbf{params(M)} & \textbf{size(MB)} \\ \hline
4DMOS & 15 & 65 & \textbf{1.84} & \textbf{7} \\
MambaMos & 3 & 321 & 184.78 & 564 \\
MF-MOS & 8 & 121 & 35.59 & 132 \\
MotionBEV & 34 & 29 & 4.42 & 17 \\
KDMOS & \textbf{40} & \textbf{25} & 4.08 & 15 \\ \hline
\end{tabular}
\end{table}

To evaluate the generalization ability of the proposed WDCD, we applied it to other MOS methods \cite{2,6,7,8}, using them as baseline models and training from scratch. The experimental results are shown in Fig.~\ref{fig:6}. The proposed module also enhances the performance of other MOS models. Notably, due to the nature of logits-based knowledge distillation, it introduces no additional parameters while improving the model's performance without any loss.
Furthermore, to explore the advantages brought by our method,we compared WDCD with other KD algorithms\cite{12,14,sun2024logit,hao2024one}, As shown in Tab.~\ref{tab:8}. Our WDCD demonstrates superior performance over other methods in MOS task.

To further validate our argument in Section~\ref{1}, we conduct an ablation study on each distillation module for non-moving classes using the SemanticKITTI-MOS validation set, while applying WDCD to the moving class as usual. The results are shown in Table~\ref{tab:tckd&nckd}, where TCKD and NCKD represent the distillation of the teacher and student models on the target and non-target classes, respectively.Since the number of non-moving classes is significantly higher than that of moving classes, they achieve higher accuracy during training. As a result, TCKD has a limited effect and may even be detrimental (-0.3\% IoU). In contrast, applying NCKD alone proves more effective than combining both.

\subsection{Qualitative Analysis}

To provide a more intuitive comparison between our algorithm and other SoTA algorithms, we conducted a visual qualitative analysis on the SemanticKITTI-MOS dataset. As shown in Fig.~\ref{fig:5}, both MF-MOS and MotionBEV exhibit misclassification of movable objects and missed detection of moving objects. Compared to SoTA algorithms, the knowledge distillation-based model effectively mitigates the impact of moving targets and accurately captures them.
\subsection{Evaluation of Resource Consumption}
We evaluate the inference time (FPS, ms), memory usage (size), and the number of learnable parameters (params) of our method and SoTA methods on Sequence 08, using 112 Intel(R) Xeon(R) Gold 6330 CPUs @ 2.00GHz and a single NVIDIA RTX 4090 GPU. As shown in Tab.~\ref{tab:performance}. We achieved real-time processing speed and outperformed MotionBEV in  terms of performance, achieving a balance between accuracy and inference speed.

\section{CONCLUSIONS}

This paper presents an efficient knowledge distillation framework tailored for MOS, enabling effective transfer of knowledge from a large-parameter teacher model to a smaller student model. The introduction of Dysample reduces model complexity and overfitting risks. Experimental results show that: 1) The proposed KDMOS model achieves high accuracy and fast inference on the SemanticKITTI-MOS dataset, balancing accuracy and speed; 2) The framework exhibits strong performance and generalization, improving various MOS methods without loss.

\bibliographystyle{IEEEtran}
\bibliography{reference}

\end{document}